\pdfoutput=1

\documentclass[11pt]{article}

\usepackage[preprint]{acl}

\usepackage{times}
\usepackage{latexsym}
\usepackage[T1]{fontenc}
\usepackage[utf8]{inputenc}

\usepackage{microtype}
\usepackage{inconsolata}
\usepackage{graphicx}
\usepackage{hyperref}
\usepackage{booktabs}
\usepackage{subcaption}
\usepackage{multirow}
\usepackage{makecell}
\usepackage{amsmath}
\usepackage{amssymb}
\usepackage{xcolor}
\usepackage{xspace}
\usepackage{tikz}
\usepackage{enumitem,amssymb}
\usepackage{wrapfig}
\usepackage{soul}

\setlist[itemize]{leftmargin=5.5mm}

\usepackage[most]{tcolorbox}
\usepackage[framemethod=tikz]{mdframed}
\usepackage{listings}
\definecolor{mybrown}{RGB}{128,64,0}
\definecolor{red}{RGB}{235, 37, 64}
\definecolor{blue}{RGB}{64, 123, 211}

\newcommand{\rev}[1]{{{#1}}}

\newcommand\blfootnote[1]{%
  \begingroup
  \renewcommand\thefootnote{}\footnote{#1}%
  \addtocounter{footnote}{-1}%
  \endgroup
}

\title{Bridging the Gap between Expert and Language Models: \\
 Concept-guided Chess Commentary Generation and Evaluation}

\author{
 \textbf{Jaechang Kim\textsuperscript{1}},
 \textbf{Jinmin Goh\textsuperscript{2,3}},
 \textbf{Inseok Hwang\textsuperscript{1,2}},
 \textbf{Jaewoong Cho\textsuperscript{4}},
 \textbf{Jungseul Ok\textsuperscript{1,2}}
\\
\\
 \textsuperscript{1}Graduate School of AI, POSTECH \\
 \textsuperscript{2}Department of Computer Science and Engineering, POSTECH \\
 \textsuperscript{3}Department of Chemical Engineering, POSTECH \\
 \textsuperscript{4}KRAFTON
}

\begin{document}
\maketitle
\begin{abstract}

Deep learning-based expert models have reached superhuman performance in decision-making domains such as chess and Go. However, it is under-explored to explain or comment on given decisions although it is important for model explainability and human education. The outputs of expert models are accurate, but yet difficult to interpret for humans.  On the other hand, large language models (LLMs) can produce fluent commentary but are prone to hallucinations due to their limited decision-making capabilities. To bridge this gap between expert models and LLMs, we focus on chess commentary as a representative task of explaining complex decision-making processes through language and address both the generation and evaluation of commentary. We introduce Concept-guided Chess Commentary generation (\texttt{CCC}) for producing commentary and GPT-based Chess Commentary Evaluation (\texttt{GCC-Eval}) for assessing it. \texttt{CCC} integrates the decision-making strengths of expert models with the linguistic fluency of LLMs through prioritized, concept-based explanations. \texttt{GCC-Eval} leverages expert knowledge to evaluate chess commentary based on informativeness and linguistic quality. Experimental results, validated by both human judges and \texttt{GCC-Eval}, demonstrate that \texttt{CCC} generates commentary which is accurate, informative, and fluent.
 \blfootnote{ \hspace{-5mm}
 \textbf{Correspondence to:} Jungseul Ok (\href{mailto:jungseul@postech.ac.kr}{jungseul@postech.ac.kr})
 }

\end{abstract}

\section{Introduction}

\begin{figure}
    \centering
    \includegraphics[width=\linewidth]{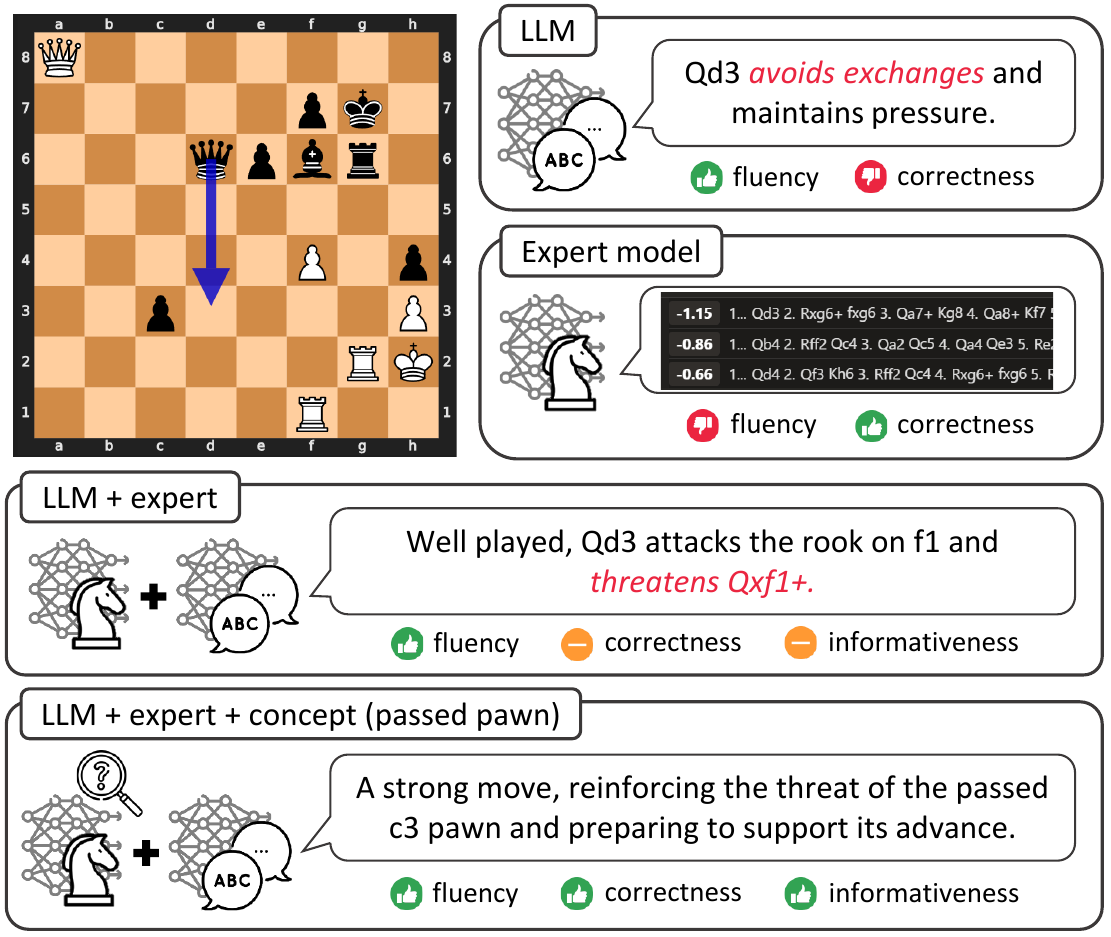}
    \caption{
        \emph{Comparison of chess commentary generation methods.}
        The \emph{\color{red}red color} indicates incorrect information.
    }
    \vspace{-5mm}
    \label{fig:limitation_prior_work}
\end{figure}


Artificial intelligence (AI) has achieved superhuman performance in various decision-making tasks, particularly in abstract strategy games like chess and Go. Milestones such as Deep Blue’s victory over the world chess champion \cite{deepblue} and AlphaGo’s defeat of top human Go players highlight AI’s capabilities in solving complex problems \cite{alphago}. While these expert models deliver highly accurate decisions, they often lack interpretability, which is critical for human education and trust in AI systems. The strategic insights and rationales behind decisions are often explained through natural language commentary~\cite{chernev2003logical, polgar2014judit}. Large language models (LLMs) exhibit their outstanding performance in generating fluent natural language. However, LLMs often struggle with hallucinations due to their limited capability in complex decision-making and lack of domain-specific knowledge.

We aim to bridge the gap between expert and language models. Specifically, we focus on the task of chess commentary generation to explain given decisions. Although chess is a resourceful testbed with extensive dataset and study \cite{zang19scc, lee2022meta, feng2023chessgpt}, the chess commentary generation has two main challenges: (i) producing accurate and insightful commentary, which requires deep chess knowledge and linguistic ability, and (ii) developing evaluation metrics to assess commentary quality, which is overlooked in previous research.

Although language models can generate fluent natural language, they lack the chess-specific knowledge required for chess commentary generation. Even a model~\cite{feng2023chessgpt} trained on chess-related data struggles in reasoning and understanding complex positions. One promising approach is to integrate expert models with language models. However, prior attempts~\cite{zang19scc, lee2022meta} 
directly feeding the decision-making process of expert models to language models 
are inadequate because the decision-making process is hard to interpret for language models. 

To address them, we introduce an effective approach using concept-based explanations of expert models. By extracting and prioritizing concepts that the expert model focuses on, we guide the language model to concentrate on the most important aspects of the game. 
This results in commentary that is both linguistically fluent and strategically insightful.
Figure~\ref{fig:limitation_prior_work} illustrates previous approaches and our approach.
Our experiments demonstrate that our approach achieves human-level correctness in commentary generation, while outperforming baselines and human-generated comments in informativeness (relevance, completeness) and linguistic quality (clarity, fluency). 

Evaluating chess commentary generation is another challenge task. Previous works~\cite{jhamtani18gac, zang19scc, lee2022meta} rely on similarity-based metrics such as BLEU, which are insufficient due to the inherently diverse nature of commentary. Different commentators may focus on distinct aspects of a position, such as attack strategies or defensive plans. 
In tasks like summarization or translation, which share the same challenges, LLM-based evaluation metrics~\cite{unieval, geval} are proposed to assess multiple dimensions.
We adopt \texttt{G-Eval}~\cite{geval} by incorporating expert model guidance for chess knowledge. 
We measure the commentary's informativeness (relevance, completeness) and linguistic quality (clarity, fluency). 
Through our experiments, we show that our proposed method correlates well with human judgments, offering a more reliable metric for commentary evaluation.

Our contributions are as follows: 
\vspace{-\topsep}
\begin{itemize} 
  \setlength{\parskip}{4pt}
  \setlength{\itemsep}{0pt plus 1pt}
\item We propose an approach that integrates expert models with LLMs through concept-based explanations, facilitating transparent decision-making in chess commentary generation. 
\item We develop a prioritization mechanism that highlights important concepts and an LLM inference technique that enables the model to understand moves with concept guidance.
\item We introduce and validate an LLM-based evaluation metric to assess the quality of chess commentary across multiple dimensions. 
\end{itemize}

\section{Related work}

\paragraph{Chess commentary generation}

Chess commentary generation is generating a comment for a chess move.
\citet{jhamtani18gac} first address the task by utilizing web-crawled data to form a chess commentary dataset, framing commentary generation as a sequence prediction problem. Building on this, \citet{zang19scc} incorporate domain-specific chess knowledge using internal chess models, improving quality and contextual relevance of generated comments.
\citet{lee2022meta}, integrate BART~\cite{bart} and an external chess engine for more reliable move evaluation. 
However, their system classifies moves into predefined categories (e.g., excellent, good, inaccuracy, mistake, blunder), without deeper understanding of the model decision-making process. In contrast, we leverage concept-based explanation to extract chess concepts from an expert model to understand the rationale behind the decision.

Not limited to chess commentary, \citet{feng2023chessgpt} fine-tune an LLM on chess-related data, to leverage chess skills, not only the linguistic ability. 
However, we demonstrate that its understanding of chess knowledge is inferior to \texttt{GPT-4o}~\cite{GPT-4} (Section~\ref{subsec:llm_chess_skill_eval}).

\paragraph{Concept-based explanation in chess}
Concepts are high-level abstractions commonly shared within a community, enabling efficient communication. In chess, concepts such as "king safety" (i.e., all potential threats against the king) condense complex strategies into understandable terms, allowing players to communicate effectively without lengthy explanations. These concepts are understandable to both humans and language models, serving as a bridge between human intuition and neural networks. Concept-based explanations aim to make a model interpretable by aligning its internal decision-making process with these shared concepts, assuming that such concepts are linearly embedded in the representation space~\cite{TCAV, alain2016understanding, mcgrath2022acquisition}. This assumption is validated in chess domains~\cite{ijcai2023stockfish, mcgrath2022acquisition} for chess expert models like \texttt{Stockfish}~\cite{stockfish}, \texttt{AlphaZero}~\cite{alphazero}, and their open-source versions, such as \texttt{LeelaChessZero}~\cite{lc0}.

\paragraph{Prioritization of concepts}
\citet{yuksekgonul23posthoc} train a post-hoc concept bottleneck model, and the classifier following the concept bottleneck model is directly interpreted as the global importance of concepts for a class.
However, they focus on finding global concept importance per class, without addressing the varying significance of concepts for individual inputs.
We address prioritization of concepts for individual inputs, or local importance, to determine the influence of each concept in specific situations.

\paragraph{Evaluation of natural language generation}

Classical evaluation metrics for natural language generation (NLG) are based on similarity.
Common metrics are BLEU~\cite{papineni2002bleu} and ROUGE~\cite{lin2004rouge}. 
However, these metrics fail to assess content quality~\cite{reiter2009investigation} and syntactic
correctness~\cite{stent2005evaluating}, and are insufficient to measure the reliability of NLG systems.
\citet{bertscore, zhao2019moverscore} compare the similarity in the text embedding space, to adequately measure semantic similarity.

Recently, beyond the similarity, \citet{yuan2021bartscore, mehri2020usr} assess generated natural language in multiple dimensions, and 
\citet{unieval, geval} evaluate in multiple dimensions using language models.
The idea of using LLMs for evaluation is common, and the evaluation methods are known to be aligned with human evaluation, sometimes more than agreements among human evaluators~\cite{rafailov2024direct, chen2023models}.
The LLM-based evaluators are focused on summarization and translation tasks. 
Regarding evaluation in chess commentary, they still lack the domain-specific knowledge required for evaluating chess commentary. 

Evaluating chess commentary is challenging due to its diverse nature, where commentaries on the same move may vary significantly depending on the focus, such as attack strategies, defensive plans, or comparison with other moves. 
Chess knowledge is essential for evaluating the correctness and relevance of these commentaries.
Previous chess commentary researches~\cite{jhamtani18gac, zang19scc, lee2022meta} use classical metrics such as BLEU, ROUGE, or perplexity, but these metrics fall short for chess commentary, as they do not evaluate with domain-specific knowledge.
While manual evaluation by human experts remains ideal, 
we propose an automatic evaluation method leveraging an LLM with chess knowledge.

\begin{figure*}[t]
    \centering
    \resizebox{0.9\linewidth}{!}{%
    \includegraphics[width=\linewidth]{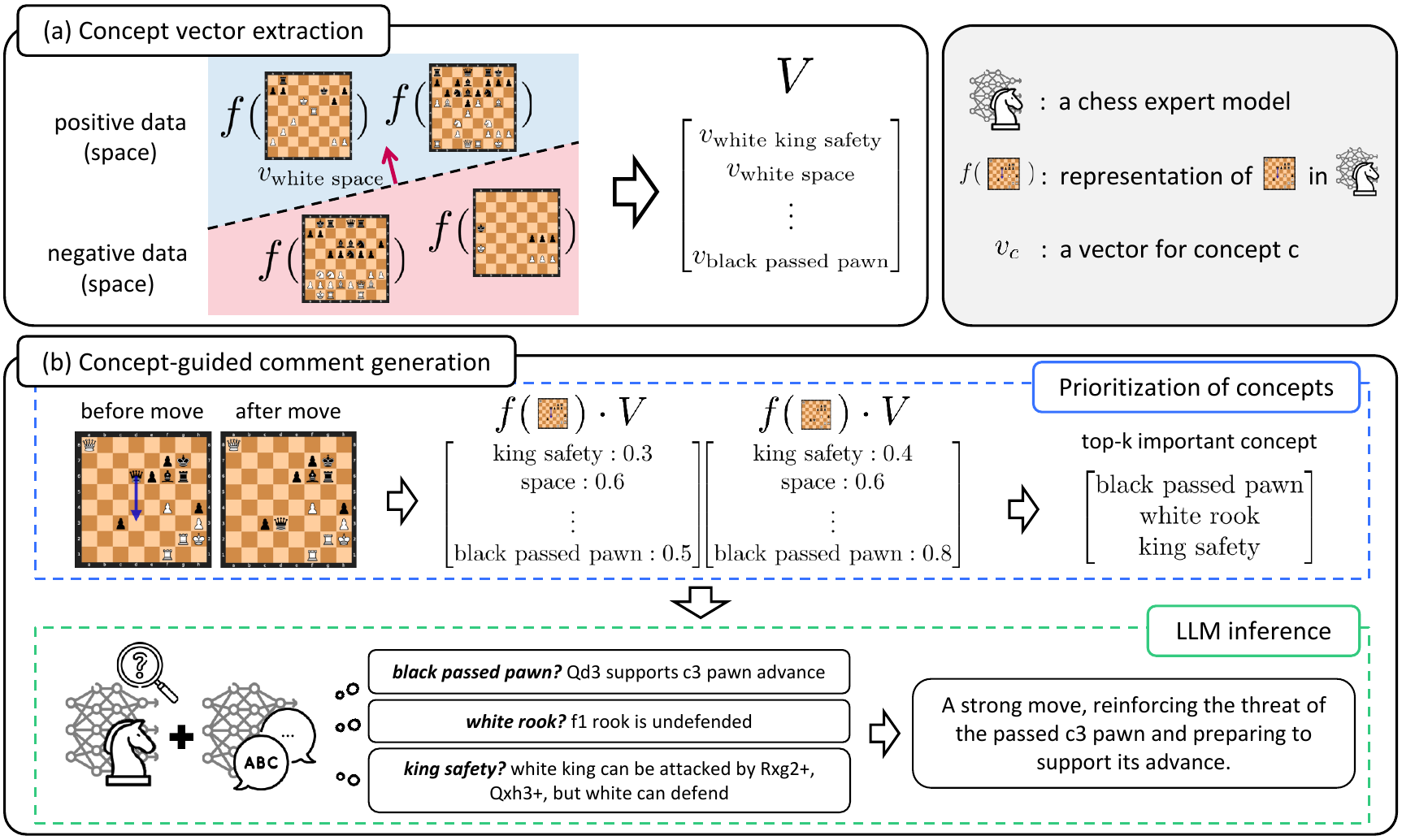}
    }
    \vspace{-1mm}
    \caption{
        \emph{Overview of \texttt{CCC},
        consists of (a) extracting concept vectors and (b) generating concept-guided commentary.
        }
    }
    \vspace{-9mm}
    \label{fig:method_overview}
\end{figure*}

\section{Method: generation and evaluation}

We propose two methods to address chess commentary generation (Section~\ref{subsec:chess_comment_generation}) and chess commentary evaluation (Section~\ref{subsec:chess_comment_evaluation}).

\subsection{Concept-guided commentary generation}
\label{subsec:chess_comment_generation}

We propose Concept-guided Chess Commentary generation (\texttt{CCC}), which is a method for generating chess commentary by leveraging a chess expert model and its concept-based explanations. 
The method involves two key steps: 
1) extracting concept vectors from a chess expert model (Section~\ref{subsec:concept_explanation}); and
2) generating commentary via an LLM using prioritized concepts that explain the given position and movement (Section~\ref{subsec:comment_generation}). 
Figure~\ref{fig:method_overview} provides an overview of the proposed method.
 
\subsubsection{Concept vector extraction}
\label{subsec:concept_explanation}
To make a chess expert model interpretable, we extract concept vectors that correspond to key concepts in chess. 
We follow a common approach~\cite{TCAV, yuksekgonul23posthoc} involving two steps: preparing a dataset for concept learning and extracting concept vectors by training a linear classifier. The concepts we focus on are adopted from \texttt{Stockfish 8}, a classical chess engine that can evaluate positions for their relevance to specific concepts (see Table~\ref{tab:concept_result}). 
We collect 200,000 chess positions from the Lichess open database~\footnote{\url{https://database.lichess.org/\#evals}} and use \texttt{Stockfish 8} to assign a score reflecting how strongly each position relates to these concepts. We then label the top $5\%$ of positions with the highest scores as positive samples and the bottom $5\%$ with the lowest scores as negative samples. This process results in a dataset of 20,000 positions for each concept, split equally between positive and negative samples. We employ \texttt{LeelaChessZero T78}, an open-source neural network-based chess model similar to \texttt{AlphaZero} for extracting concept vectors. For the representation space, we use the final layer before policy and value heads (layer 40). We then train a linear Support Vector Machine (SVM)~\citep{svm} to classify these samples. The resulting normal vector of the SVM classification boundary serves as the concept vector, and the distance from this boundary determines the concept score for any input position. This score quantifies how strongly a given board state aligns with the extracted concept.

\subsubsection{Chess comment generation with an expert model and extracted concepts}
\label{subsec:comment_generation}

\paragraph{Prioritization of concepts} 
Given a chess position and a specific move, our goal is to identify the concepts most relevant to explaining that movement.
For the chess position, we compute the score for each concept by taking the dot product between the expert model representation of the position and the extracted concept vectors. These concept scores reflect how strongly each concept is reflected in the current position.
To prioritize concepts, we compare the concept scores before and after the move. By analyzing the differences between pre-move and post-move scores, we identify which concepts are most influenced by the move. This allows us to assign priority to the concepts that explain the impact of the move.



\begin{figure*}[t]
    \centering
    
    \begin{tcolorbox}[colback=mybrown!5!white,colframe=mybrown!75!black]
    \begin{small}
    \begin{minipage}[t]{0.25\linewidth}
    \vspace*{0pt}
        \includegraphics[height=\linewidth,width=\linewidth]{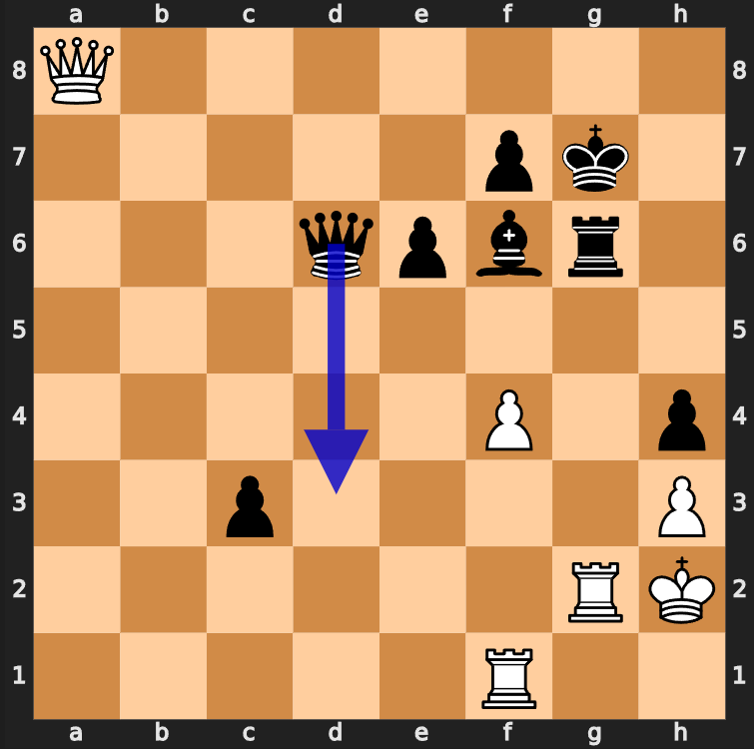}
    \end{minipage}
    \hspace{5pt}
    \begin{minipage}[t]{0.73\linewidth}
    \vspace*{0pt}
        \begin{itemize}[leftmargin=2mm]
        \setlength\itemsep{0em}
        \item \textbf{Reference comment} \\
            Exchanging rook will give more oportunities to attack the king and find the mate.
        \item \textbf{\texttt{GAC}~\cite{jhamtani18gac}} \\
            Black moves his queen to a safe square.
        \item \textbf{\texttt{GPT-4o}} \\
            Qd3 is a solid move, \emph{\color{red}avoiding exchanges} and maintaining pressure. Black's queen is now targeting several critical squares and creating threats. White needs to find accurate defensive moves to stay in the game.
        \item \textbf{\texttt{GPT-4o + expert}} \\
            well played, Qd3 attacks the rook on f1 and \emph{\color{red}threatens Qxf1+}. White now faces a tough defense, expected move is g2-g6 to create counter-play.
        \item \textbf{\texttt{GPT-4o + expert + concept}}
            (Pawns, {\color{blue}\ul{Black Passedpawns}}, White Kingsafety)\\
            Qd3 is a strong move, reinforcing the threat of the {\color{blue}\ul{passed c3 pawn  and preparing to support its advance}}. It also keeps pressure on White's position, maintaining the initiative.
        \end{itemize}
    \end{minipage}\hfill%
    \end{small}
    \end{tcolorbox}
    \vspace{-2mm}
    \caption{
        \emph{Examples of generated comments.} 
        \emph{\color{red} Red text} denotes incorrect information, and {\color{blue}\ul{blue text}} denotes important concepts and affected counterparts.
    }
    \label{fig:comment_examples}
\end{figure*}

\paragraph{Commentary generation via LLM}
We generate chess commentary using an LLM and a chess expert model.
Although a language model understands chess-specific notations and terms, it lacks the ability to perform chess-specific reasoning and complex analysis, which can result in hallucination.
By integrating chess expert model output, the LLM determines whether to focus on advantageous aspects or disadvantageous aspects.
However, since the chess expert model output is based on scalar values, it still generates incorrect comments.
Concept-based explanation guides the LLM to focus on critical aspects.
Figure~\ref{fig:comment_examples} is a typical example of a concept-guided comments.

To enhance the reasoning ability of LLM, we employ few-shot prompting, Chain-of-Thought (CoT) prompting~\cite{wei2022chain}, and chess-specific information.
This approach provides the LLM with a deeper understanding of chess positions, and prevents potential use of wrongly prioritized concepts.
Additionally, we enumerate all existing attacks towards opponent pieces to prevent mentioning of non-existing pieces or illegal moves.

\subsection{Automatic evaluation of commentary}
\label{subsec:chess_comment_evaluation}
\label{sec:auto_eval_chess_comment}

Our evaluation approach, termed \texttt{GCC-Eval}, modifies and extends \texttt{G-Eval} to better address the specific challenges of evaluating chess commentary. 
The core components of \texttt{GCC-Eval} are:
(i) Multi-dimensional evaluation by an LLM.
(ii) Expert model evaluation for chess knowledge.
(iii) Auto-CoT for score-only output.
(iv) weighted summation for non-integer scores.
Note that our contributions are on the first and second aspects to ensure accurate chess commentary evaluation, focusing on informativeness and linguistic quality.

\paragraph{Evaluation dimensions}
The evaluation covers four dimensions: relevance, completeness, clarity, and fluency. 
While clarity and fluency are general linguistic measures, relevance and completeness require a deep understanding of chess. 
To address this, we employ an expert model to augment the LLM’s capabilities when scoring relevance and completeness. 
This integration ensures that the commentary is not only linguistically sound but also informative from a domain-expert perspective.

\rev{
LLM generates a single-integer evaluation between one to five after Auto-CoT.
The probability of each score represents the confidence of LLM for the score.
We use a weighted summation of score probabilities as follows:
}
\begin{align} \mathrm{score}(x) = \sum_{s \in \{1,2,3,4,5\}}{s \times p(s | x)} . \end{align}
This method allows to produce non-integer scores in LLM-based evaluation, capturing subtle nuances in the evaluation that would be missed by integer-only scoring schemes.
The scoring prompts, including the expert evaluation and Auto-CoT reasoning, are described in Appendix~\ref{sec:appendix_G_eval_cc}.




\begin{table*}[t]
\vspace{-3mm}
    \centering
    \resizebox{\textwidth}{!}{
        \begin{tabular}{ ccccccc } 
            \toprule
            \textbf{Comment generation methods} &  \makecell{\textbf{Correctness}} & \makecell{\textbf{Relevance} } & \makecell{\textbf{Completeness}} & \makecell{\textbf{Clarity}} & \makecell{\textbf{Fluency}} & \makecell{\textbf{Words per}\\\textbf{comment}} \\ 
            \midrule
            \midrule
\multicolumn{1}{l}{Reference} &0.62	& $^*$0.52	& $^*$0.30	& $^*$0.60	& $^*$0.62 & $^*$15.6 \\ 
            \midrule
            \multicolumn{1}{l}{\texttt{GAC}~\cite{jhamtani18gac}} &\textbf{0.63}	& $^*$0.46	&$^*$0.15	&$^*$0.66	&$^*$0.64 &$^*$8.9 \\ 
            \multicolumn{1}{l}{\texttt{GPT-4o}} &$^*$0.36	&$^*$0.49	&$^*$0.40	&\underline{$^*$0.72}	&$^*$0.84 & $^*$27.1 \\ 
            \multicolumn{1}{l}{\texttt{GPT-4o + expert}} &$^*$0.43	&\underline{$^*$0.56}	&\underline{$^*$0.49}	&\underline{$^*$0.72}		&\underline{$^*$0.85} &$^*$26.2 \\ 
            \multicolumn{1}{l}{\texttt{GPT-4o + expert + concept} (\texttt{CCC}, ours)} &\underline{0.60} &\textbf{0.67}	&\textbf{0.59}	&\textbf{0.80}	&\textbf{0.91} &28.5 \\ 
            \bottomrule
        \end{tabular}
    }
    \caption{
        \emph{Average scores of human evaluation.}
        Bold and underlined text indicate the best and second-best methods in each column, respectively.
        Numbers are rescaled to the range $[0, 1]$.
        \rev{
        $^*$ denotes the difference from \texttt{CCC} is significant (i.e., p-value $< 0.001$).
        }
    }
    \label{tab:human_eval}
\end{table*}

\begin{table*}[t]
\vspace{-2mm}
    \centering
    \resizebox{0.9\linewidth}{!}{
    \begin{tabular}{ cccccc } 
        \toprule
        \textbf{Types of errors} & \texttt{{GPT-4o}} & \makecell{\texttt{{GPT-4o}} \\ \texttt{{+ expert}}} & \makecell{\texttt{{GPT-4o}} \\ \texttt{{+ expert + concept}}} \\ 
        \midrule
        \midrule
        \makecell[l]{Referring illegal move or non-existing pieces} & 0.46 & 0.28 & 0.20 \\ 
        \makecell[l]{Wrong understanding of tactical/immediate advantage} & 0.46 & 0.40 & 0.26 \\ 
        \makecell[l]{Wrong understanding of positional/long-term advantage} & 0.28 & 0.26 & 0.28 \\ 
        \makecell[l]{Wrong evaluation of the move/position} & 0.32 & 0.30 & 0.34 \\ 
        \bottomrule
    \end{tabular}%
    }
    \caption{
        \emph{Error rates in different causes of incorrectness.} 
        Note that the questions allow multiple answers per question.
        Error types in the lower rows require more comprehensive reasoning.
    }
    \vspace{-3mm}
    \label{tab:error_types}
\end{table*}

\begin{table}[t]
    \centering
    \resizebox{0.5\linewidth}{!}{
        \begin{tabular}{ cc } 
            \toprule
            \textbf{Metrics} & \makecell{\textbf{$\kappa$}} \\ 
            \midrule
            \midrule
            \multicolumn{1}{l}{Correctness} & 0.5393 \\
            \multicolumn{1}{l}{Relevance} & 0.2448 \\
            \multicolumn{1}{l}{Completeness} & 0.2449 \\
            \multicolumn{1}{l}{Clarity} & 0.1782 \\
            \multicolumn{1}{l}{Fluency} & 0.2328 \\
            \bottomrule
        \end{tabular}
    }
    \caption{
        \emph{Inter-annotator agreements of human evaluation measured by Fleiss' kappa($\kappa$).}
    }
    \vspace{-5mm}
    \label{tab:rank_fleiss_kappa}
\end{table}

\setul{1pt}{.4pt}

\begin{table*}[t]
    \centering
    \resizebox{0.75\linewidth}{!}{%
        \begin{tabular}{ ccccccccccc } 
            \toprule
            \multirow{2}{*}{\textbf{Metrics}} & \multicolumn{2}{c}{\textbf{Correctness}} & \multicolumn{2}{c}{\textbf{Relevance}} & \multicolumn{2}{c}{\textbf{Completeness}} & \multicolumn{2}{c}{\textbf{Clarity}} & \multicolumn{2}{c}{\textbf{Fluency}} \\ 
             & $\rho$ & $\tau$&  $\rho$ & $\tau$&  $\rho$ & $\tau$&  $\rho$ & $\tau$&  $\rho$ & $\tau$ \\
            \midrule
            \midrule
            \multicolumn{1}{l}{BLEU-1} &  0.17 & 0.10 & -0.06  & 0.02 & -0.25 & -0.07 & -0.08 & 0.01 & -0.36 & -0.16  \\ 
            \multicolumn{1}{l}{ROUGE-1} & 0.04 & -0.01 & -0.19 & -0.08 & -0.29 & -0.18 & -0.18 & -0.14 & -0.29 & -0.17 \\ 
            \multicolumn{1}{l}{ROUGE-2} & 0.15 & 0.03 & -0.10 & -0.03 & -0.16 & -0.05 & -0.02 & 0.03 & -0.15 & 0.00  \\ 
            \multicolumn{1}{l}{ROUGE-L} & 0.08 & 0.01 & -0.18 & -0.08 & -0.29 & -0.18 & -0.16 & -0.14 & -0.29 & -0.17 \\ 
            \midrule
            \multicolumn{1}{l}{\texttt{GCC-Eval}} & - & - & 0.40 & 0.24 & 0.56 & 0.39 & 0.44 & 0.23 & 0.55 & 0.38 \\ 
            \bottomrule
        \end{tabular}%
    }
    \caption{
    \emph{Correlations between human and automatic evaluations.}
    $\rho$ and $\tau$ denotes Pearson correlation and Kendall's tau correlation, respectively.
    }
    \vspace{-3mm}
    \label{tab:human_auto_correlation}
\end{table*}

\begin{table*}[t]
    \centering
     \resizebox{0.9\textwidth}{!}{%
        \begin{tabular}{ cccccc } 
            \toprule
            \textbf{Comment generation methods} & \makecell{\textbf{Relevance} } & \makecell{\textbf{Completeness}} & \makecell{\textbf{Clarity}} &  \makecell{\textbf{Fluency}} \\ 
            \midrule
            \midrule
            \multicolumn{1}{l}{Reference} & 0.51 & 0.25 & 0.47  & 0.72 \\ 
            \midrule
            \multicolumn{1}{l}{\texttt{GAC}~\cite{jhamtani18gac}} & 0.47 & 0.14 & 0.39  & 0.81 \\ 
            \multicolumn{1}{l}{\texttt{GPT-4o}} & 0.79 & 0.48 & 0.85  & 0.95 \\ 
            \multicolumn{1}{l}{\texttt{GPT-4o + expert}} & 0.81 & 0.49 & 0.75  & 0.90 \\ 
            \multicolumn{1}{l}{\texttt{GPT-4o + expert + concept} (\texttt{CCC}, ours)} & \textbf{0.89} & \textbf{0.54} & \textbf{0.88}  & \textbf{1.00} \\ 
            \bottomrule
        \end{tabular}%
     }
    \caption{
        \emph{Automatic evaluation results using \texttt{GCC-Eval}.}
        Numbers are rescaled to the range $[0, 1]$.
    }
    \vspace{-3mm}
    \label{tab:auto_eval}
\end{table*}

\begin{table}[t]
    \centering
    \resizebox{\linewidth}{!}{%
    \begin{tabular}{ ccccc } 
        \toprule
        \multirow{1}{*}{\textbf{Concepts}} & \textbf{Accuracy} & \textbf{Precision} & \textbf{Recall} \\
        \midrule
        \midrule
        \multicolumn{1}{l}{Material}  & 0.93 & 0.93 & 0.94   \\
        \multicolumn{1}{l}{Imbalance}  & 0.80 & 0.73    & 0.93   \\
        \multicolumn{1}{l}{Pawns}  & 0.84 & 0.81    & 0.90   \\
        \multicolumn{1}{l}{White Knights}  & 0.91 & 0.87    & 0.96   \\
        \multicolumn{1}{l}{Black Knights}  & 0.91 & 0.87    & 0.97   \\
        \multicolumn{1}{l}{White Bishop}  & 0.77 & 0.73 & 0.87   \\
        \multicolumn{1}{l}{Black Bishop}  & 0.75 & 0.71 & 0.83   \\
        \multicolumn{1}{l}{White Rooks}  & 1.00 & 1.00  & 1.00   \\
        \multicolumn{1}{l}{Black Rooks}  & 0.99 & 0.99  & 1.00   \\
        \multicolumn{1}{l}{White Queens}  & 0.74 & 0.71 & 0.79   \\
        \multicolumn{1}{l}{Black Queens}  & 0.81 & 0.84 & 0.77   \\
        \multicolumn{1}{l}{White Mobility}  & 0.99 & 0.99   & 1.00   \\
        \multicolumn{1}{l}{Black Mobility}  & 0.98 & 0.96   & 0.99   \\
        \multicolumn{1}{l}{White Kingsafety}  & 0.96 & 0.97 & 0.94   \\
        \multicolumn{1}{l}{Black Kingsafety}  & 0.94 & 0.96 & 0.91   \\
        \multicolumn{1}{l}{White Threats}  & 0.93 & 0.90    & 0.96   \\
        \multicolumn{1}{l}{Black Threats}  & 0.93 & 0.90    & 0.97   \\
        \multicolumn{1}{l}{White Space}  & 1.00 & 1.00  & 1.00   \\
        \multicolumn{1}{l}{Black Space}  & 1.00 & 1.00  & 1.00   \\
        \multicolumn{1}{l}{White Passedpawns}  & 0.98 & 0.98    & 0.98   \\
        \multicolumn{1}{l}{Black Passedpawns}  & 0.92 & 0.91    & 0.94   \\
        \bottomrule
    \end{tabular}
    }
    \caption{
    \emph{Test accuracy, precision and recall of concept-based explanations.}
    }
    \vspace{-5mm}
    \label{tab:concept_result}
\end{table}

\section{Experiments}
\label{sec:experiments}

\subsection{Experimental settings}

\paragraph{Dataset} 
We evaluate our model using \emph{Chess Commentary} dataset introduced by \citet{jhamtani18gac}. 
This dataset contains full chess games accompanied by user-generated commentary on specific moves, collected from an online chess forum\footnote{\url{https://gameknot.com/}}. Following the train/valid/test split introduced by \citet{jhamtani18gac}, we use only the test set for our experiments. Since the absence of pre-processing code, we manually align the raw data with pre-processed data to ensure fair comparison with \texttt{GAC}. 
Additionally, we exclude comments that covering multiple moves for simplicity in analysis.

\paragraph{Baselines} 
We compare the experimental results within several methods: 
\vspace{-\topsep}
\begin{itemize} 
  \setlength{\parskip}{4pt}
  \setlength{\itemsep}{0pt plus 1pt}
    \item \textbf{reference}: These are reference texts from the \texttt{GameKnot} dataset.
    \item \textbf{\texttt{GAC}}~\cite{jhamtani18gac}: An LSTM model trained on the \texttt{GameKnot} dataset for generating chess commentary. 
    \item \textbf{\texttt{GPT-4o}}~\cite{GPT-4}: The unmodified version of \texttt{GPT-4o}, accessed via OpenAI API, with a temperature setting of $0.1$ to avoid noisy outputs. 
    For detailed discussion of comparison of LLMs, refer to Section~\ref{subsec:llm_chess_skill_eval}.
    \item \textbf{\texttt{GPT-4o + expert}}: This is the same \texttt{GPT-4o} model but augmented with evaluations from a chess expert model. 
    Note that \citet{lee2022meta} use BART with a chess expert model and \texttt{GPT-4o + expert} is superior because it uses more powerful language model and a sufficient expert model.
\end{itemize}

\paragraph{Reproducibility}
\rev{
For better reproducibility, we release our source code~\footnote{\url{https://github.com/ml-postech/concept-guided-chess-commentary}}.
}

\subsection{Human evaluation}
\label{subsec:exp_human_eval}

\paragraph{Human evaluation settings}
\rev{
We conducted a manual human evaluation to assess the quality of the generated chess commentary. All participants were fluent in English, both in reading and writing, and possessed sufficient chess knowledge to evaluate the comments reliably. Specifically, we recruited five participants from the university community and social networks. Each participant held a Chess.com rapid rating above 1500, placing them within the top 0.49\% among chess players\footnote{retrieved in Oct 2024, from \url{https://www.chess.com/leaderboard/live/rapid}}. The average rating among participants was 1776. Participants were compensated with an amount equivalent to 73 USD, which matched the local minimum wage for 10 hours of work. The evaluation plan was approved by our university's Institutional Review Board (IRB number: PIRB-2024-E028). All participants were informed about the purpose of the research and the intended use of the evaluation data, and consent was obtained before participation. The evaluation results were anonymized and did not contain any personal information. To enhance the quality of the evaluation process, we conducted two rounds of pilot tests with two participants who were not involved in the main evaluation. These pilot studies helped us clarify potentially ambiguous expressions, adjust the response format, estimate the time required, and optimize the timing of mandatory breaks based on participant feedback.

The evaluation followed a within-participant design. For each move, participants evaluated commentary generated by five methods (four baselines and \texttt{CCC}), with the order of methods randomized. A total of 50 moves were evaluated, resulting in 250 comments assessed by each participant. The evaluation took approximately four hours to complete, including breaks. Participants were required to take breaks after every 50 comments (i.e., approximately 10-minute breaks after 40 minutes of evaluation) to ensure sustained focus.
During the evaluation, participants were presented with a chessboard displaying a specific move, marked with a blue arrow. Alongside the moves, the corresponding commentaries were provided.
Each participant was asked to rate the commentary across six questions: five evaluation metrics and one question for categorizing the type of incorrectness when applicable. The evaluated metrics were correctness, relevance, completeness, clarity, and fluency.
Relevance and completeness assessed how informative and insightful the comment was, while clarity and fluency evaluated its linguistic naturalness.
Relevance, completeness, clarity, and fluency were assessed using a five-point Likert scale, while correctness was evaluated using a three-point Likert scale, as the correctness of a comment was closer to a binary decision rather than a scaled question.
For clear presentation, the scores were rescaled to a range of 0 to 1.
Appendix~\ref{sec:appendix_human_eval} provides the instructions and questions used for human evaluation.
}


\paragraph{Main results}
Table~\ref{tab:human_eval} presents the results of the human evaluation. Our proposed method, \texttt{CCC}, achieves the highest scores in all metrics except correctness, where it ranks second. 
Also, \texttt{CCC} outperforms the reference comments in every metric except correctness, and the correctness is also comparable to the reference. 
The reference comments, collected from online sources, often contain grammatical mistakes and informal language, underscoring the limitation of similarity-based evaluation metrics. This highlights the need for evaluation metrics beyond similarity, especially when the quality of the reference comments is suboptimal.
The use of expert models and concept guidance contribute significantly to the overall performance improvement, as evidenced by the higher scores across most metrics. 
While \texttt{GPT-4o + expert} shows only a slight improvement in correctness, it generates more detailed explanations, which in some cases lead to minor factual inaccuracies in the details, as illustrated in Figure~\ref{fig:comment_examples}. 
Although \texttt{GAC} exhibits the highest correctness slightly outperforming \texttt{CCC}, we observe that \texttt{GAC}'s higher correctness comes at the cost of lower details of the explanations; their explanations tend to be brief and thereby less informative in general, leading to lower scores in completeness and shorter the comment lengths.


\paragraph{Detailed analysis}
Table~\ref{tab:error_types} provides a detailed analysis of the types of errors. 
The usage of the expert model and concept reduces simple errors, but errors requiring comprehensive understanding remain within the margin of error.

To validate the consistency of the human evaluation, we calculate inter-annotator agreement using Fleiss' Kappa~\cite{fleiss_kappa} of ranks across different methods. Table~\ref{tab:rank_fleiss_kappa} reports the agreement of the participants. The agreement for correctness is 0.54, indicating moderate agreement. This is notably higher than for other metrics, suggesting that correctness is more indisputable for chess experts, compared to more subjective qualities like relevance or fluency.

\subsection{Automatic evaluation}
\label{subsec:exp_auto_eval}

To perform an automatic evaluation of generated chess commentaries, we employ our proposed metric, \texttt{GCC-Eval}. This metric is designed to assess both linguistic quality and domain-specific relevance in chess commentary. To validate its reliability, we calculate the correlation between \texttt{GCC-Eval} scores and human evaluations using the same dataset from prior human evaluation studies. As shown in Table~\ref{tab:human_auto_correlation}, \texttt{GCC-Eval} consistently shows a higher correlation with human assessments across all evaluation criteria compared to traditional metrics, such as BLEU and ROUGE, which rely on surface-level similarity measures with reference comments.
We further apply \texttt{GCC-Eval} to evaluate the performance of different chess commentary generation methods. The results in Table~\ref{tab:auto_eval} indicate that \texttt{CCC} outperforms the baselines in all \texttt{GCC-Eval} metrics, showcasing the effectiveness of integrating domain-specific expertise and concept-based explanations.

\subsection{Other experiments}

\paragraph{Chess skills and knowledge of language model}
\label{subsec:llm_chess_skill_eval}

While LLMs can generate linguistically sound commentary, they lack the deep, inherent understanding of chess strategies. Integrating expert models like chess engines compensates for this limitation, ensuring that the LLM's output is both fluent and grounded in expert knowledge.
To verify the chess skill level of LLMs, we use mate-in-one chess problems and evaluate how the models solve them, in Table~\ref{tab:mate_puzzle}.
\texttt{GPT-4o} solves 57\% of problems, while other language models are below 12\%, even though \texttt{ChessGPT}~\cite{feng2023chessgpt} is fine-tuned on chess-related documents.
When the expert model evaluation result is given in prompt, the LLM solves 95\% of the problems, which is not surprising because the expert model evaluation includes the answer.
While \texttt{GPT-4o + expert} includes the answer in the prompt, \texttt{GPT-4o + concept} also shows significant improvement of 17.2\%p, with only a simple hint that there is a mate.
It implies that a proper concept serves as a powerful hint for the precise analysis of the position.
For more detailed explanation, refer to Appendix~\ref{sec:appendix_chess_skill}.

\paragraph{Reliability of the concept-based explanation}
We assess the reliability of the concept-based explanations. 
Table~\ref{tab:concept_result} shows that the average accuracy of the extracted chess concepts is 0.91, demonstrating that the model effectively identifies and utilizes key domain-specific concepts. This further supports the idea that concept-based explanations serve as reliable source for guiding the LLM in generating chess comments.

\paragraph{Interactive commentary generation}

We also explore the potential of \texttt{CCC} for generating interactive and context-aware chess commentary. By augmenting the LLM with the decision-making capabilities of an expert model, it responds to flexible user questions, providing deeper insights beyond simple commentary on a move. 
The questions can be strategic intentions, long-term plans, and potential threats in a given chess position. An example of these interactive commentary capabilities and corresponding results are found in Appendix~\ref{sec:appendix_interactive}. 
These experiments demonstrate that \texttt{CCC} is capable of generating not only accurate move annotations, but also high-quality interactive chess insights that meet different requirements of different users.

\begin{table}[t]
    \centering
    \resizebox{\linewidth}{!}{
        \begin{tabular}{ cccc } 
            \toprule
            \textbf{Language models} & \textbf{LLM} & \textbf{\makecell{LLM \\ + expert}} & \textbf{\makecell{LLM \\ + concept\\(mate-in-one)}}\\
            \midrule
            \midrule
            \multicolumn{1}{l}{\texttt{GPT-4o}} & 0.564 & 0.982 & 0.736 \\
            \multicolumn{1}{l}{\texttt{GPT-4o-mini}} & 0.014 & 0.988 & 0.031 \\
            \multicolumn{1}{l}{\texttt{GPT-3.5-turbo}} & 0.036 & 0.988 & 0.056 \\
            \multicolumn{1}{l}{{\texttt{ChessGPT} \cite{feng2023chessgpt}}} & 0.118 & 0.563 & 0.175 \\
            \bottomrule
        \end{tabular}
    }
    \caption{
    \emph{LLM chess skill evaluation on mate-in-one problems.}
    }
    \vspace{-3mm}
    \label{tab:mate_puzzle}
\end{table}

\section{Discussions}
\label{sec:discussion}

\paragraph{Language model as an explanation form}

Our work shows that the \texttt{CCC} framework effectively transfers AI-driven chess knowledge to human users.
Beyond concept-based explanation, language models can act as a crucial medium between the expert model's internal reasoning and the end-user.
This connection facilitates more intuitive and understandable feedback than traditional explanation methods like saliency-based, which suffer from issues of inconsistency and unreliability.
By employing language-based form of explanation, the transparency of the explanation can be improved, making the evaluation of the model's reliability more straightforward.

\paragraph{Fine-tuning with \texttt{GCC-Eval}}

We validate that \texttt{GCC-Eval} is well-correlated with human evaluation.
One promising direction to improve the quality of chess commentary is to incorporate \texttt{GCC-Eval} as a training objective, replacing human evaluator.
By optimizing models to directly align with this evaluative criterion, we can better ensure that the generated commentary meets the standards of human chess experts. 
This approach offers a potential pathway toward more robust and human-aligned commentary systems in future applications.

\rev{
\paragraph{Applicability of \texttt{CCC}}

While we focus on chess in this paper as a first step, our approach is applicable to other domains, such as science~\cite{tshitoyan2019unsupervised, sprague2018interpretable}, medical imaging~\cite{yeche2019ubs}, and games\cite{das2023state2explanation} where concept-based explanation methods have been actively studied. 
well-established concept extractors and relevant datasets in those domains make the domains suitable for applying our method.

\paragraph{Results with PGN chess notation}

We also verify that \texttt{gpt-4o} understands PGN notation, not only FEN notation.
Inferring from \texttt{CCC} results with PGN notation, the differences from that with FEN notation are not noticible, but we cannot validate it in large scale because our dataset does not contain the game progresses which is required for using PGN notation.
Appendix~\ref{sec:appendix_pgn} provides the results with PGN notation.
}

\section{Conclusions}

In this paper, we propose methods for chess commentary generation (\texttt{CCC}) and evaluation (\texttt{GCC-Eval}). 
\texttt{CCC} integrates expert and language models through concept-based explanations, utilizing techniques such as prioritization, few-shot learning, and Chain-of-Thought prompting to align effectively with expert knowledge. 
\texttt{CCC} either surpasses or matches the quality of human-generated commentary, demonstrating the capability of LLMs to express expert-level understanding and potentially enhance learning for human users.
We also present \texttt{GCC-Eval}, a multi-dimensional evaluation framework that incorporates chess-specific knowledge to assess chess commentary. 
The strong correlation between human evaluation and \texttt{GCC-Eval} validates the robustness. 
These findings underscore promising future research directions, including using a language model as an explanation method and 
using \texttt{GCC-Eval} fine-tuning chess commentary generation models.

\section{Limitations}

\paragraph{Use of proprietary LLMs}
\rev{
We release the source code used in our experiments. 
}
However, since we employed proprietary LLMs including \texttt{GPT-4o} (from July to October 2024), it can be limited to fully reproduce the results. Nonetheless, the proposed framework remains adaptable and can be further enhanced with the integration of more advanced LLMs. In addition, it is also interesting to further investigate the efficacy of our framework with smaller LLMs.

\paragraph{Educational purpose / comment for beginners}
The main audience for commentary is often beginners and or those with less knowledge than the commentator. In the human evaluation in Section~\ref{subsec:exp_human_eval}, we assess the commentary in the view of expert chess players.
Another human evaluation involving novice players can assess the educational impact of the comments.
For the same purpose, \citet{chen2023models} propose counterfactual simulatability, as an automatic evaluation metric of the improvement of students.

\paragraph{Beyond chess commentary}
Although we focus on the chess commentary generation, our method can be extended to other tasks, that require comprehensive decision-making abilities and have an expert model.
Empirical experiments in other tasks require finding the appropriate tasks and corresponding expert models.

\paragraph{More concepts}
Although we use concepts from \texttt{Stockfish 8}, there are other useful concepts such as fork, pin, double-pawn or open-file.
We do not use the concepts because of insufficient concept labels, but they could be valuable, as the concept "mate-in-one" improves chess skill in Table~\ref{tab:mate_puzzle}.

\paragraph{Differences between concept evaluation function and extracted concept}

In our work, we extract the concept vectors from an expert model. 
Although using oracle concept evaluation functions is relatively more accurate, there are two key reasons for using the extracted concepts.
First, recent findings~\cite{schut2023bridging} emphasize that expert models often possess super-human knowledge, capturing patterns and strategies not easily interpretable by humans. It implies the extracted concepts can cover the comprehensive knowledege of model, even if the humans do not understand and an oracle concept evaluation function is not present.
Second, when the model has defects, the extracted concepts are used to find the cause of failure.
These two aspects facilitate us to use extracted concepts.

\section*{Acknowledgement}

\rev{

This work was supported by Institute of Information \& communications Technology Planning \& Evaluation (IITP) grant funded by the Korea government Ministry of Science and ICT (MSIT) (RS-2019-II191906, Artificial Intelligence Graduate School Program (POSTECH); IITP-2024-RS-2024-00437866; RS-2024-00509258, Global AI Frontier Lab), and a grant from KRAFTON AI.

}

\bibliography{custom}

\clearpage 
\appendix

\renewcommand{\thefigure}{A\arabic{figure}}
\setcounter{figure}{0}

\label{sec:appendix}

\section*{Appendix}

\section{Details for \texttt{GCC-Eval}}
\label{sec:appendix_G_eval_cc}

\begin{figure*}[t]
    \centering
    \begin{subfigure}[]{\textwidth}
        \centering
        \begin{tcolorbox}[colback=mybrown!5!white,colframe=mybrown!75!black]
        \begin{small}
        \textbf{[System]}\\
        You will be given single comment about a chess move.\\
        Your task is to rate the comment on one metric.\\
        Please make sure you read and understand these instructions carefully. Please keep this document open while reviewing, and refer to it as needed.\\
        \\
        Evaluation Criteria:\\
        Relevance (1-5) - Relevence of a target comment. The comment should include only information relevant to the chess move or reasoning for taking or not taking the chess move. An engine evaluation result is given as a hint.\\
        Evaluation Steps:\\
        1. Read the comment carefully.\\
        2. Assess how well the comment addresses the important information about the chess move, and how relevant it is.\\
        3. Assign a Relevance score from 1 to 5.\\
        \\
        \textbf{[User]}\\
        position: \emph{\color{blue} 8/3nk3/1p4pp/1N1P1p2/1bP2KP1/3P1P2/7P/8 b - - 0 0}\\
        move: \emph{\color{blue}30... Bd2+}\\
        target comment: \emph{\color{blue}Good move, Bd2+ forces the White king to move, gaining tempo and improving the position of the Black bishop.}\\
        engine evaluation: \emph{\color{blue} actual move - Bd2+ 232cp, expected reply - f4g3, best move - Bd2+ similar to actual move, second best move - Nc5 similar to actual move}\\
        Score(1-5, score ONLY):
        \end{small}
        \end{tcolorbox}
        \caption{Example prompt of relevance.}
    \end{subfigure}
    \begin{subfigure}[]{\textwidth}
        \centering
        \begin{tcolorbox}[colback=mybrown!5!white,colframe=mybrown!75!black]
        \begin{small}
        \textbf{[System]}\\
        You will be given single comment about a chess move.\\
        Your task is to rate the comment on one metric.\\
        Please make sure you read and understand these instructions carefully. Please keep this document open while reviewing, and refer to it as needed.\\
        \\
        Evaluation Criteria:\\
        Completeness (1-5) - Completeness of a comment. The comment should cover all critical points on the chess board, ensuring that no important factors are overlooked. An engine evaluation result is given as a hint.\\
        Evaluation Steps:\\
        1. Read the comment carefully.\\
        2. Assess how well the comment addresses the important information, and how well the comment covers the entire important information without missing any.\\
        3. Assign a Completeness score from 1 to 5.\\
        \\
        \textbf{[User]}\\
        position: \emph{\color{blue} 8/3nk3/1p4pp/1N1P1p2/1bP2KP1/3P1P2/7P/8 b - - 0 0}\\
        move: \emph{\color{blue} 30... Bd2+}\\
        target comment: \emph{\color{blue} Good move, Bd2+ forces the White king to move, gaining tempo and improving the position of the Black bishop.}\\
        engine evaluation: \emph{\color{blue} actual move - Bd2+ 232cp, expected reply - f4g3, best move - Bd2+ similar to actual move, second best move - Nc5 similar to actual move}\\
        Score(1-5, score ONLY):
        \end{small}
        \end{tcolorbox}
        \caption{Example prompt of completeness.}
    \end{subfigure}
\end{figure*}
\begin{figure*}[t]\ContinuedFloat
    \centering
    \begin{subfigure}[]{\textwidth}
        \centering
        \begin{tcolorbox}[colback=mybrown!5!white,colframe=mybrown!75!black]
        \begin{small}
        \textbf{[System]}\\
        You will be given single comment about a chess move.\\
        Your task is to rate the comment on one metric.\\
        Please make sure you read and understand these instructions carefully. Please keep this document open while reviewing, and refer to it as needed.\\
        \\
        Evaluation Criteria:\\
        Clarity (1-5) - Clarity of a comment. The comment should be clear and detailed, without vague or ambiguous statements.\\
        Evaluation Steps:\\
        1. Read the commment carefully.\\
        2. Assess how the comment is clear and detailed, without vague or ambiguous statements.\\
        3. Assign a Clarity score from 1 to 5.\\
        \\
        \textbf{[User]}\\
        position: \emph{\color{blue} 8/3nk3/1p4pp/1N1P1p2/1bP2KP1/3P1P2/7P/8 b - - 0 0}\\
        move: \emph{\color{blue} 30... Bd2+}\\
        comment: \emph{\color{blue} Good move, Bd2+ forces the White king to move, gaining tempo and improving the position of the Black bishop.}\\
        Score(1-5, score ONLY):
        \end{small}
        \end{tcolorbox}
        \caption{Example prompt of clarity.}
    \end{subfigure}
    \begin{subfigure}[]{\textwidth}
        \centering
        \begin{tcolorbox}[colback=mybrown!5!white,colframe=mybrown!75!black]
        \begin{small}
        \textbf{[System]}\\
        You will be given one comment written for a chess move.\\
        Your task is to rate the comment on one metric.\\
        Please make sure you read and understand these instructions carefully. Please keep this document open while reviewing, and refer to it as needed.\\
        \\
        Evaluation Criteria:\\
        Fluency (1-5): Fluency of a comment.\\
        1. Read the commment carefully.\\
        2. Assess the sentences of comment is coherently organized. The comment should contain well-structured language and coherent transitions.\\
        3. Assign a Fluency score from 1 (not readable) to 5 (very fluent).\\
        \\
        \textbf{[User]}\\
        target comment: \emph{\color{blue} Good move, Bd2+ forces the White king to move, gaining tempo and improving the position of the Black bishop.}\\
        Score(1-5, score ONLY):
        \end{small}
        \end{tcolorbox}
        \caption{Example prompt of fluency.}
    \end{subfigure}
        \caption{
        \emph{Example prompts for \texttt{GCC-Eval}.}
        The \emph{\color{blue} blue text} in the figure changes according to the experimental conditions.
        }
    \label{fig:prompt_g_eval_cc}
\end{figure*}

The scoring prompt, which includes the expert evaluation and Auto-CoT reasoning, is illustrated in Figure~\ref{fig:prompt_g_eval_cc}.

\section{Reproduction of baselines}

Although \citet{jhamtani18gac} provide the source code, necessary pre-processing files are missing.
For fair comparison, we align the raw files with the pre-processed files to compare reference text and chess move to the generated comments.
For the same reason, we cannot reproduce \citet{zang19scc}.
\citet{lee2022meta} do not share the source code, and as our baseline \texttt{GPT-4o + expert} shares the same idea with it, we do not reproduce it.

\section{Human evaluation examples}
\label{sec:appendix_human_eval}

\begin{figure*}[t]
    \centering

    \begin{tcolorbox}[colback=mybrown!5!white,colframe=mybrown!75!black]
    \begin{small}
    \textbf{Fluency:} \\
    You are asked to indicate your level of agreement to that ``The commentary flows smoothly, with well-structured language and coherent transitions between sentences.''\\
    \\
    Example of fluent comment:  A bold pawn push, h5 aims to challenge white's center and possibly prepare for further expansion on the kingside.  \\
    Example of not fluent comment:  A pawn push, h5 to white's Center and expansion the kingside.\\
    \\
    This diagram compares relevance and completeness. High relevance indicates C is small, and high completeness indicates A is small. \\
    \\
    \frame{\includegraphics[width=0.7\columnwidth]{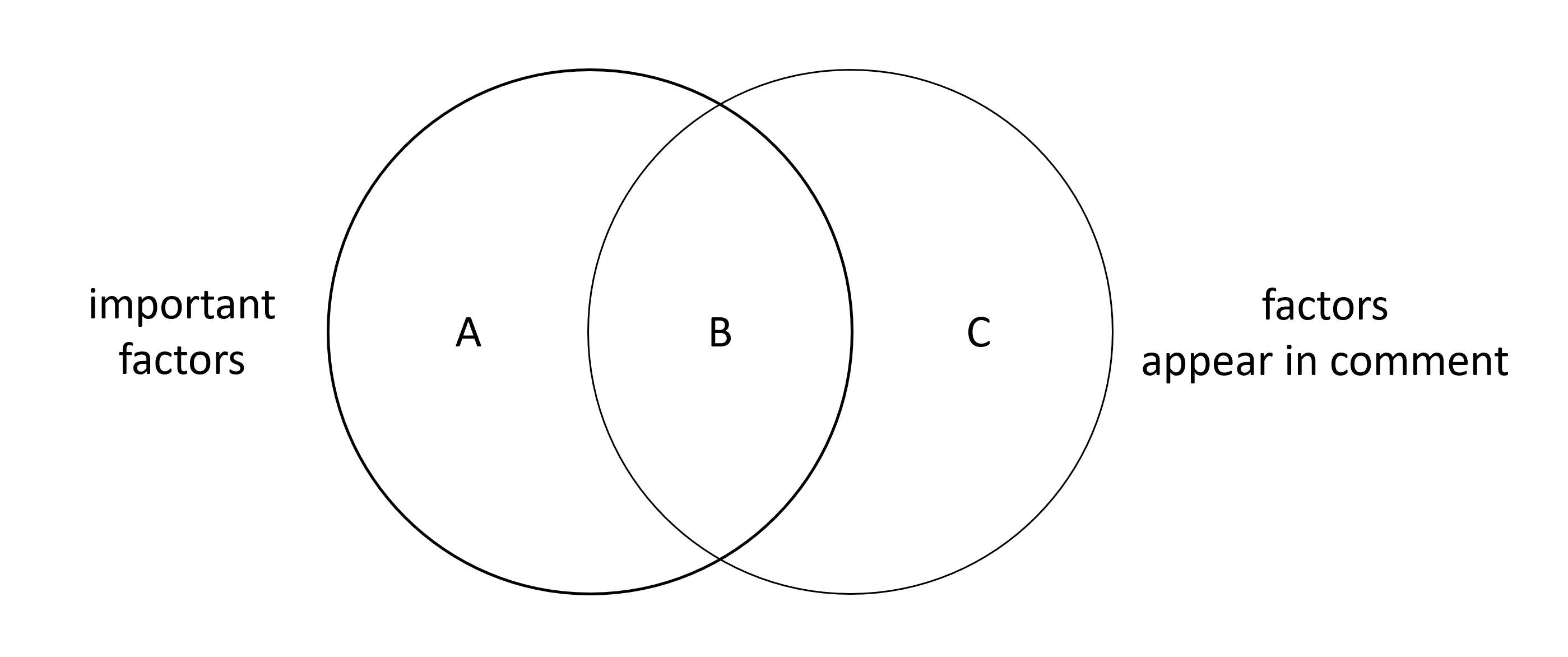}}
    \end{small}
    \end{tcolorbox}
    \caption{
    \emph{Survey instruction.}
    }
    \label{fig:human_eval_form_instruction}
\end{figure*}

Figure~\ref{fig:human_eval_form_instruction} and 
Figure~\ref{fig:human_eval_form_example} are instructions and questions we used for human evaluation.

\begin{figure*}[t]
    \centering
    
    \begin{tcolorbox}[colback=mybrown!5!white,colframe=mybrown!75!black]
    \begin{small}
    \textbf{Chess Commentary Evaluation: Instruction}\\ 
    \\
    \textbf{About survey}\\
    This survey asks you to evaluate comments on chess moves. For each move (shown in blue arrow on a chessboard), five different comments for the same move are provided. For each comment, you are asked to evaluate it based on five metrics: \textbf{correctness, relevance, completeness, clarity, and fluency}. In what follows, we explain each metric and then provide example evaluations on three comments.\\
    \\
    Please do not spend too much time analyzing the chessboard. \textbf{Each page is expected to be completed in one  minute (although you may need some extra time at the beginning)}.\\
    \\
    \includegraphics[width=0.3\columnwidth]{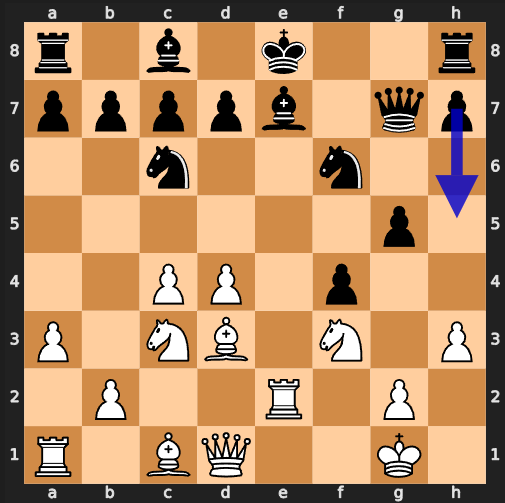}\\
    \\
    \textbf{Explanation of each criteria}\\
    Please refer to the image above and read example comments with it.\\
    \\
    \textbf{Correctness:}\\
    You are asked to indicate your level of agreement to that ``The commentary provides accurate analysis, ensuring that all evaluations and moves are logically and factually correct.''\\
    \\
    Example of correct sentence: Black expands kingside with h5.\\
    Example of incorrect sentence (misuse of move/pieces): Black attacks knight.\\
    Example of incorrect sentence (wrong understanding/description of tactical advantage): White can fork knight and bishop in the next move.\\
    Example of incorrect sentence (wrong understanding/description of positional / long-term advantage): Black's kingside castling will improve king safety.\\
    \\
    \textbf{Relevance: }\\
    You are asked to indicate your level of agreement to that ``The commentary focuses on the key aspects of the move or game without including unnecessary or unrelated information.''\\
    \\
    Example of relevant sentence: Black expands kingside with h5.\\
    Example of irrelevant sentence: Black's c8 bishop has not activated.\\
    \\
    \textbf{Completeness:} \\
    You are asked to indicate your level of agreement to that ``The commentary covers all critical aspects of the position, ensuring that no important factors are overlooked.''\\
    
    Example of complete comment: Black expands kingside with h5, supported by rook. But black still have weaknesses in pinned e7 bishop and c6 knight can be threatened by d5.\\
    Example of incomplete comment: Black expands kingside with h5.\\
    \\
    \textbf{Clarity:} \\
    You are asked to indicate your level of agreement to that ``The commentary is clear and detailed, without vague or ambiguous statements.''\\
    \\
    Example of clear sentence: Black is starting kingside pawn push, followed by g4 and f3.\\
    Example of unclear sentence:  Black is starting kingside pawn push.
    \end{small}
    \end{tcolorbox}
\end{figure*}

\makeatletter
\newcommand*{\radiobutton}{%
  \@ifstar{\@radiobutton0}{\@radiobutton1}%
}
\newcommand*{\@radiobutton}[1]{%
  \begin{tikzpicture}
    \pgfmathsetlengthmacro\radius{height("X")/2}
    \draw[radius=\radius] circle;
    \ifcase#1 \fill[radius=.6*\radius] circle;\fi
  \end{tikzpicture}%
}
\makeatother

\newlist{todolist}{itemize}{2}
\setlist[todolist]{label=$\square$}

\begin{figure*}[t]
    \centering

    \begin{tcolorbox}[colback=mybrown!5!white,colframe=mybrown!75!black]
    \begin{small}
    \textbf{Chess Commentary Evaluation} \\
    \\
    Position and move (1/250)\\
    \\
    \includegraphics[width=0.25\columnwidth]{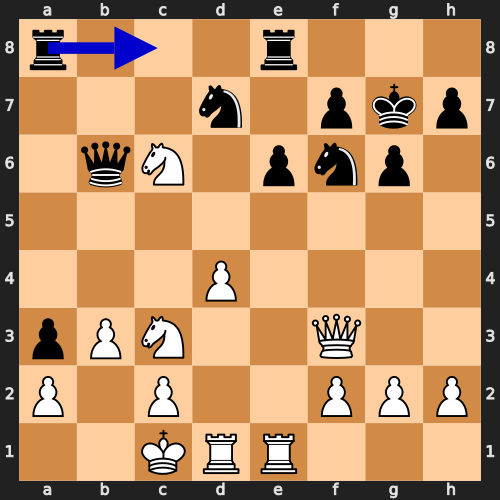}
    \\
    Comment to be evaluated:\\
    20... Rac8 is a solid move, reinforcing the c-file and preparing to challenge White's strong knight on c6. However, White still maintains a strong position with active knights and potential threats.\\
    \\
    Correctness: The commentary provides accurate analysis, ensuring that all evaluations and moves are logically and factually correct.\\
    .\quad\quad\quad\;\;\;\;\:1\quad\quad\quad\quad\quad\;2\quad\quad\quad\quad\quad\,\,3\\
    disagree \quad\radiobutton\quad\quad\quad\quad\quad\radiobutton\quad\quad\quad\quad\quad\radiobutton\quad agree\\
    \\
    If the comment is incorrect, what is the incorrect point?
    \begin{todolist}
        \item referring illegal move or non-existing pieces
        \item wrong understanding/usage of tactical/immergent advantage
        \item wrong understanding/usage of positional/long-term advantage
        \item wrong evaluation of the move
        \item etc:
    \end{todolist}
    Relevance: The commentary focuses on the key aspects of the position or game without including unnecessary or unrelated information.\\
    .\quad\quad\quad\quad\quad\quad\;\;\;\;\;\;\:1\quad\quad\quad\quad\quad\;2\quad\quad\quad\quad\quad\,\,3\quad\quad\quad\quad\quad4\quad\quad\quad\quad\quad\;5\\
    strongly disagree \quad\radiobutton\quad\quad\quad\quad\quad\radiobutton\quad\quad\quad\quad\quad\radiobutton\quad\quad\quad\quad\quad\radiobutton\quad\quad\quad\quad\quad\radiobutton\quad strongly agree\\
    \\
    Completeness: The commentary covers all critical aspects in the position, ensuring that no important factors are overlooked.\\
    .\quad\quad\quad\quad\quad\quad\;\;\;\;\;\;\:1\quad\quad\quad\quad\quad\;2\quad\quad\quad\quad\quad\,\,3\quad\quad\quad\quad\quad4\quad\quad\quad\quad\quad\;5\\
    strongly disagree \quad\radiobutton\quad\quad\quad\quad\quad\radiobutton\quad\quad\quad\quad\quad\radiobutton\quad\quad\quad\quad\quad\radiobutton\quad\quad\quad\quad\quad\radiobutton\quad strongly agree\\
    \\
    Clarity: The commentary is clear and detailed, without vague or ambiguous statements.\\
    .\quad\quad\quad\quad\quad\quad\;\;\;\;\;\;\:1\quad\quad\quad\quad\quad\;2\quad\quad\quad\quad\quad\,\,3\quad\quad\quad\quad\quad4\quad\quad\quad\quad\quad\;5\\
    strongly disagree \quad\radiobutton\quad\quad\quad\quad\quad\radiobutton\quad\quad\quad\quad\quad\radiobutton\quad\quad\quad\quad\quad\radiobutton\quad\quad\quad\quad\quad\radiobutton\quad strongly agree\\
    \\
    Fluency: The commentary flows smoothly, with well-structured language and coherent transitions between sentences.\\
    .\quad\quad\quad\quad\quad\quad\;\;\;\;\;\;\:1\quad\quad\quad\quad\quad\;2\quad\quad\quad\quad\quad\,\,3\quad\quad\quad\quad\quad4\quad\quad\quad\quad\quad\;5\\
    strongly disagree \quad\radiobutton\quad\quad\quad\quad\quad\radiobutton\quad\quad\quad\quad\quad\radiobutton\quad\quad\quad\quad\quad\radiobutton\quad\quad\quad\quad\quad\radiobutton\quad strongly agree
    \end{small}
    \end{tcolorbox}
    \caption{
    \emph{Survey example.}
    }
    \label{fig:human_eval_form_example}
\end{figure*}


\section{Interactive commentary}
\label{sec:appendix_interactive}

\begin{figure*}[t]
    \centering
    
    \begin{tcolorbox}[colback=mybrown!5!white,colframe=mybrown!75!black]
    \begin{small}
    \includegraphics[width=0.3\columnwidth]{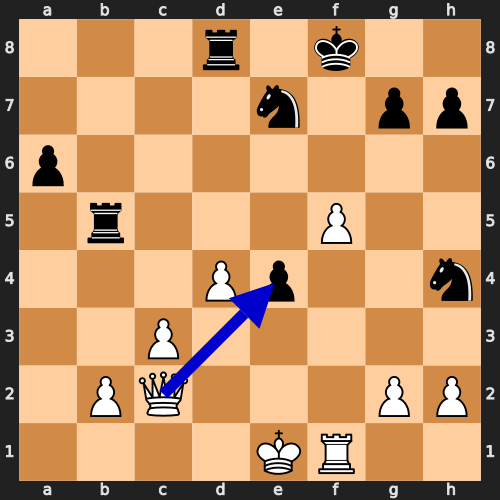}\\
    \textbf{[Comment of LLM:]}\\
    comment: Qxe4 is an even trade but misses a stronger opportunity with Qf2. While the move captures the passed e4 pawn, it exposes White to counterplay, particularly b5-b2, which could generate strong pressure on White’s queenside. Better would have been Qf2, maintaining positional advantage and limiting Black’s activity.
    \\
    \textbf{[Following question by user:]}\\
    After the move, can black's h4 knight survive?\\
    \textbf{[Response of LLM:]}\\
    After 26. Qxe4, Black's knight on h4 is under threat. The White queen can capture the knight with Qxh4. Black doesn't have any immediate way to defend or save the knight on h4 effectively.
    \\
    Given that the knight is undefended, and White can simply take it on the next move, the knight cannot survive unless Black manages to create a very strong counterattack that would force White to deal with something else first, but that seems unlikely based on the current position.
    \\
    Thus, it looks like Black's knight on h4 cannot survive and is likely lost after White's next move.
    \end{small}
    \end{tcolorbox}
    \caption{
    \emph{Example of interactive commentary.}
    }
    \label{fig:interactive_comment_example}
\end{figure*}

Figure~\ref{fig:interactive_comment_example} shows an example of interactive comments, starting from \texttt{CCC}. 
The initial chess commentary is generated by \texttt{CCC}. 
If there are parts of the generated comments that are unclear or difficult to understand, users can engage with the system by asking follow-up questions to clarify any ambiguous or complex parts of the commentary.
Similarly, they can request additional insights, such as alternative moves or a deeper analysis of the current game position.

This interactive approach enhances knowledge transfer between the AI and users, making expert-level chess understanding more accessible. By enabling two-way communication, the functionality of LLMs is extended, transforming the model from a static generator of text into an interactive learning tool that adapts to the needs and curiosity of the user. This capability promotes a more engaging and educational experience in chess commentary, expanding the role of LLMs in expert domains.



\section{Chess skill evaluation details} 
\label{sec:appendix_chess_skill}
We conduct chess skill evaluation for LLMs. We use mate-in-one puzzle data from database of Lichess (\url{https://database.lichess.org/\#puzzles}). We conduct evaluation for 1,000 puzzle data. Evaluation prompts are shown in Figure~\ref{fig:chess_mate_prompt}. For \texttt{GPT-4o + expert}, we include expert model evaluation information in the prompt (Figure~\ref{fig:chess_mate_prompt}(a)). For \texttt{GPT-4o + concept}, we provide an explanation indicating that the board is in a mate-in-one situation (Figure~\ref{fig:chess_mate_prompt}(b)). For \texttt{GPT-4o}, \texttt{GPT-4o-mini}, \texttt{GPT-3.5-turbo}, and \texttt{ChessGPT}, we use a basic prompt for evaluation (Figure~\ref{fig:chess_mate_prompt}(c)).

\begin{figure*}[t]
    \centering
    \begin{subfigure}[]{\textwidth}
        \centering
        \begin{tcolorbox}[colback=mybrown!5!white,colframe=mybrown!75!black]
        \begin{small}
        \textbf{[System]}\\
        You will be given a chess board, formatted with Forsyth-Edwards notation(FEN) string. \\
        Your task is to find the best move of this board.\\
        Please answer the best move in standard algebraic notation(SAN). \\
        \textbf{[User]}\\
        position: N6r/1p1k1ppp/2np4/b3p3/4P1b1/N1Q5/P4PPP/R3KB1R b KQ - 0 18\\
        \emph{\color{blue}engine evaluation: best move - Bxc3\#}\\
        Move(SAN formatted move only):
        \end{small}
        \end{tcolorbox}
        \caption{Example prompt of \texttt{GPT-4o + expert}.}
    \end{subfigure}
    \begin{subfigure}[]{\textwidth}
        \begin{tcolorbox}
        [colback=mybrown!5!white,colframe=mybrown!75!black]
        \begin{small}
        \textbf{[System]}\\
        You will be given a chess board, formatted with Forsyth-Edwards notation(FEN) string. \\
        Your task is to find the best move of this board. \emph{\color{blue}You can make checkmate in one move.} \\
        Please answer the best move in standard algebraic notation(SAN). \\
        \textbf{[User]}\\
        position: N6r/1p1k1ppp/2np4/b3p3/4P1b1/N1Q5/P4PPP/R3KB1R b KQ - 0 18\\
        Move(SAN formatted move only):
        \end{small}
        \end{tcolorbox}
        \caption{Example prompt of \texttt{GPT-4o + concept} "mateIn1".}
    \end{subfigure}
    \begin{subfigure}[]{\textwidth}
        \begin{tcolorbox}[colback=mybrown!5!white,colframe=mybrown!75!black]
        \begin{small}
        \textbf{[System]}\\
        You will be given a chess board, formatted with Forsyth-Edwards notation(FEN) string. \\
        Your task is to find the best move of this board.\\
        Please answer the best move in standard algebraic notation(SAN). \\
        \textbf{[User]}\\
        position: N6r/1p1k1ppp/2np4/b3p3/4P1b1/N1Q5/P4PPP/R3KB1R b KQ - 0 18\\
        Move(SAN formatted move only):
        \end{small}
        \end{tcolorbox}
        \caption{Example prompt of \texttt{GPT-4o}, \texttt{GPT-4o-mini}, \texttt{GPT-3.5-turbo}, \texttt{ChessGPT}.}
    \end{subfigure}
        \caption{
        \emph{Example prompts for chess skill evaluation with mate-in-one problems.}
        The \emph{\color{blue} blue text} in figures (a) and (b) indicates the differences from figure (c).
        }
    \label{fig:chess_mate_prompt}
\end{figure*}

\section{Results with PGN notation}
\label{sec:appendix_pgn}
\rev{
We verify that LLM and \texttt{CCC} also works with PGN notation instead of FEN notation.
Figure~\ref{fig:pgn_result} presents an example result of \texttt{CCC} with FEN and \texttt{CCC} with PGN.
There are no significant differences in the generated commentary quality or usage of concepts.
}

\begin{figure*}[t]
    \centering
        \begin{tcolorbox}[colback=mybrown!5!white,colframe=mybrown!75!black]
        \begin{small}
        \textbf{[PGN]}\\
        1. e4 Nf6 2. Nc3 d6 3. b3 g6 4. f4 Bg7 5. Nf3 O-O 6. Bc4 Re8 7. O-O b6 8. Ne2
Nxe4 9. d3 Bxa1 10. dxe4 Bg7 11. Ng3 Nc6 12. f5 Ne5 13. fxg6 hxg6 14. Nxe5 Bxe5
15. Bxf7+ Kg7 16. Bxg6 Rh8 17. Rf5 Bxf5 18. Bxf5 Qg8 19. Nh5+ Kf7 20. Qd5+ e6 \\
        \textbf{[comment generated using PGN]}\\
        The move 21. Bxe6+ is a strong continuation for White, as it captures a pawn with check and maintains the pressure on Black's position. This move forces Black to respond to the check, giving White the initiative. The move also opens up the possibility of further attacks on Black's king, which is somewhat exposed. The White bishop on e6 is well-placed, and Black's pieces are somewhat disorganized, especially with the Black queen and rooks not ideally coordinated to defend. Overall, this move helps White maintain a significant advantage. \\
        \textbf{[FEN]}\\
        r5qr/p1p2k2/1p1pp3/3QbB1N/4P3/1P6/P1P3PP/2B3K1 w - - 0 21 \\
        \textbf{[comment generated using FEN]}\\
        The move 21. Bxe6+ is a strong tactical decision by White. By capturing the pawn on e6 with check, White not only gains material but also disrupts Black's pawn structure and exposes the Black king further. This move forces Black to respond to the check, giving White the initiative. The White bishop is well-placed, and this move capitalizes on its active position. The evaluation indicates a significant advantage for White, reflecting the effectiveness of this move.
        \end{small}
        \end{tcolorbox}
    \caption{An example of \texttt{CCC} with PGN and FEN notation.}
    \label{fig:pgn_result}
\end{figure*}

\section{Licenses of artifacts}

In this study, \texttt{GPT-4o} is used in compliance with its usage policy. 
\texttt{ChessGPT} is used under the terms of the Apache-2.0 license. 
The Lichess database is used according to the Creative Commons CC0 license. 
As there are no specific license statements for \texttt{GameKnot} and \texttt{GAC}, 
we regard them as Creative Commons CC0 license.

All artifacts are used within the intended use.

\end{document}